\documentclass[letterpaper, 10 pt, journal, twoside]{ieeetran}

\IEEEoverridecommandlockouts 

\IEEEoverridecommandlockouts                              

\usepackage{graphics} 
\usepackage{epsfig} 
\usepackage{mathptmx} 
\usepackage{times} 
\usepackage{amsmath} 
\usepackage{amssymb}  
\usepackage{booktabs} 
\usepackage{multirow}
\usepackage{cleveref}
\usepackage{subcaption}
\usepackage{makecell}   

\usepackage{graphicx}  

\usepackage{capt-of}   

\usepackage[dvipsnames]{xcolor}

\usepackage{algorithm}
\usepackage{algpseudocode}

\usepackage{makecell}
\usepackage{pifont}

\newcommand{\bad}[1]{\textcolor{Bittersweet}{#1}}
\newcommand{\good}[1]{\textcolor{ForestGreen}{#1}}
\newcommand{\best}[1]{\textcolor{ForestGreen}{\underline{#1}}}

\newcommand{\overallperfimpro}{4.3\%}
\newcommand{\overallseimpro}{36.2\%}

\begin{document}

\title{
Pretraining in Actor-Critic Reinforcement Learning for Locomotion
}


\markboth{IEEE ROBOTICS AND AUTOMATION LETTERS. PREPRINT VERSION. ACCEPTED JUNE, 2026 }
{Fan \MakeLowercase{\textit{et al.}}: Pretraining in Actor-Critic Reinforcement Learning for Locomotion} 

\author{Jiale Fan$^{1}$, Andrei Cramariuc$^{2}$, Tifanny Portela$^{3}$ and Marco Hutter$^{2}$%
\thanks{Manuscript received: February 20, 2026; Revised May 22, 2026; Accepted June 19, 2026.}
\thanks{This paper was recommended for publication by Editor O. Stasse upon evaluation of the Associate Editor and Reviewers’ comments.
This work was supported by the ETH AI Center and the European Union’s Horizon Europe Framework Programme under grant agreement No 101121321.} 
\thanks{$^{1} $Jiale Fan is with Robotic Systems Lab, ETH Zurich, 8092 Zürich, Switzerland, and also with Section of Microengineering, EPFL, 1015 Lausanne, Switzerland.
        }%
\thanks{$^{2}$Andrei Cramariuc and Marco Hutter are with  Robotic Systems Lab, ETH Zurich, 8092 Zürich, Switzerland. 
        {\tt\footnotesize jialfan@ethz.ch}}%
\thanks{$^{3} $Tifanny Portela is with Robotic Systems Lab, ETH Zurich, 8092 Zürich, Switzerland, and also with ETH AI Center, ETH Zurich, 8092 Zürich, Switzerland.
        }%

\thanks{© 2026 IEEE. Personal use of this material is permitted.
Permission from IEEE must be obtained for all other uses, in any current
or future media, including reprinting/republishing this material for
advertising or promotional purposes, creating new collective works, for
resale or redistribution to servers or lists, or reuse of any copyrighted
component of this work in other works.}%
}




\maketitle


\begin{abstract}
The pretraining-finetuning paradigm has facilitated numerous transformative advancements in artificial intelligence research in recent years. However, in the domain of reinforcement learning (RL) for robot locomotion, individual skills are often learned from scratch despite the high likelihood that some generalizable knowledge is shared across all task-specific policies belonging to the same robot embodiment. This work aims to define a paradigm for pretraining neural network models that encapsulate such knowledge and can subsequently serve as a basis for warm-starting the RL process in classic actor-critic algorithms, such as Proximal Policy Optimization (PPO). We begin with a task-agnostic exploration-based data collection algorithm to gather diverse, dynamic transition data, which is then used to train a Proprioceptive Inverse Dynamics Model (PIDM) through supervised learning. The pretrained weights are then loaded into both the actor and critic networks to warm-start the policy optimization of actual tasks. We systematically validated our proposed method with 9 distinct robot locomotion RL environments comprising 3 different robot embodiments, showing significant benefits of this initialization strategy. Our proposed approach on average improves sample efficiency by \overallseimpro \space and task performance by \overallperfimpro \space compared to random initialization. We further present key ablation studies and empirical analyses that shed light on the mechanisms behind the effectiveness of this method.
\end{abstract}


\section{Introduction}

\IEEEPARstart{T}{he} pretraining-finetuning paradigm has enabled recent major breakthroughs in computer vision \cite{he2022masked, lu2019vilbert} and natural language processing \cite{devlin-etal-2019-bert}, most notably in the case of large language models \cite{touvron2023llama, achiam2023gpt}. In the domain of robotics, a similar methodology with pre-initialization and fine-tuning has been explored in several works that integrate visual-language model (VLM) backbones for manipulation tasks \cite{black2024pi_0, team2024octo, team2025gemini, barreiros2025careful}. However, these works only address the pretraining of the vision or language backbones, which have well-studied benefits and strategies, but do not endow robots with information about embodiment. While these imitation learning-based approaches offer good generalization to different tasks, they suffer from low-frequency execution and are primarily demonstrated on stable platforms and environments, rather than on dynamically unstable robotic platforms or under substantial external disturbances.

\begin{figure*}[t!]
\vspace{2em}
    \centering
    \includegraphics[width=0.7\linewidth]{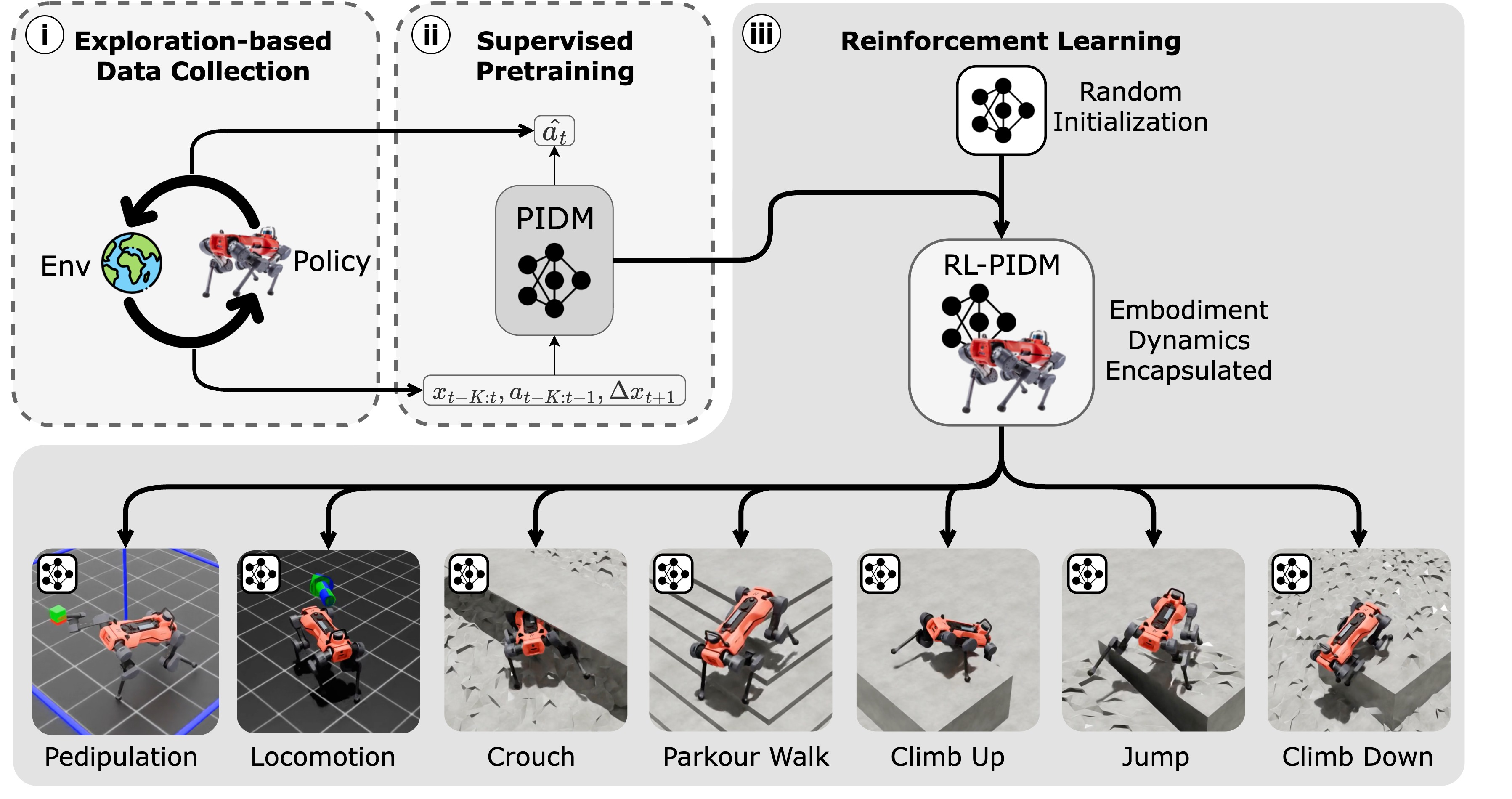}
    \caption{\textbf{Method overview:} We (i) collect task-agnostic data using an exploration-driven policy, (ii) to train a Proprioceptive Inverse Dynamics Model (PIDM) to capture embodiment-aware dynamics, and (iii) initialize the actor-critic networks in PPO to warm-start the RL process.}
    \label{fig:system_overview}
\end{figure*}

In robot locomotion control, reinforcement learning (RL) with Proximal Policy Optimization (PPO) \cite{schulman2017proximal} has been used to successfully achieve a wide range of robust and agile motions \cite{miki_learning_2022, zhang2025distillationpponoveltwostagereinforcement, yang2023neuralvolumetricmemoryvisual}. However, skill acquisition is slow and resource-intensive because RL is generally sample-inefficient and each new task is typically learned \textit{tabula rasa}, even within the same embodiment. Looking back at model-based control paradigms \cite{full_mpc_sleiman, Murphy2012HighDD}, for a specific robot embodiment, there is knowledge that is sharable across solutions to different tasks, \textit{e.g.}, the joint kinematics and dynamics of the model. Motivated by this, we posit that warm-starting RL training in actor-critic architectures by incorporating such embodiment-aware knowledge into the initial model weights has the potential to improve policy performance and sample efficiency (Fig. \ref{fig:system_overview}).

In short, this work presents a method for smart network initialization in the context of learning robot locomotion with PPO, which outperforms the commonly used random initialization~\cite{he2015delving} across various tasks with the same embodiment. Our perspective on the problem is novel as we propose a task-agnostic approach that focuses solely on encapsulating embodiment-specific knowledge across tasks. It does not need reward signal of the task-specific downstream MDPs to be present in the pretraining dataset, and serves as a user-friendly plug-in that does not require modifications to the established paradigm of locomotion learning. 
Our proposed method consists of three stages: exploration-based data collection, pretraining, and reinforcement learning. We first employ an exploration-based data collection strategy to systematically investigate states most likely to appear in the initial stages of the RL process, where the robot learns fundamental concepts about its embodiment, including limb kinematics, dynamics, and basic stability. With the collected data, we then train an embodiment-aware Proprioceptive Inverse Dynamics Model (PIDM). Finally, by initializing the actor-critic structure with the weights of the PIDM model, we provide the RL process with general knowledge from the initial stumbling stages of the vanilla training process, thus facilitating training. Our pretrained weights do not contain task-specific biases, but let them emerge naturally during RL training, as the entire network is updated in an end-to-end fashion. 

We validate this approach with a diverse locomotion skill set and multiple robot embodiments consisting of two quadrupeds and one humanoid. Our approach improves performance by \overallperfimpro \space and sample efficiency by \overallseimpro. The main contributions are:
\begin{enumerate}
    \item A paradigm of embodiment-specific weight initialization for robot locomotion RL that improves performance and sample efficiency. The initialization obtained this way is task-agnostic, \textit{i.e.}, applicable to various downstream Partially Observable Markov Decision Process (POMDP) formulations involving different commands, observations, rewards, curricula and terrains, as long as the same robot embodiment is retained.
    \item Extensive empirical validation of our proposed approach with various embodiments and tasks showcases significant improvements in performance and sample efficiency.
    \item Empirical evidence reveals a strong correlation between PIDM accuracy and downstream RL performance and suggests inverse dynamics prediction encapsulates representations relevant to locomotion policy learning.
\end{enumerate}

\section{Related Works}

\subsection{Pretraining in Reinforcement Learning}
Although RL excels on well-specified tasks, its limited sample efficiency remains a key challenge \cite{jin2021bellmaneluderdimensionnew}. Xie et al. \cite{xie_pretraining_2022} systematically summarized the efforts made to introduce the pretraining paradigm to RL.
A large body of prior work \cite{ball_efficient_2023, hansen_td-mpc2_2024, nakamoto_cal-ql_2023} shares a similar motivation with ours in seeking to improve online locomotion learning through pretraining on offline datasets. However, these methods typically rely on reward-labeled data tailored to a specific downstream task, whereas our goal is to learn task-agnostic weight initializations that generalize across multiple downstream tasks for a given embodiment. The unknowness and diversity of downstream MDPs preclude the inclusion of task-specific reward signals in the pretraining dataset.
More closely aligned with our approach are methods that learn transferable representations from unlabeled offline data. 

\subsection{Pretraining representations in RL without reward labels}  Schwarzer et al. \cite{schwarzer_pretraining_2021} employ a combination of latent dynamics modelling and unsupervised goal-conditioned RL to pretrain useful representations that can be later fine-tuned to task-specific rewards. Allen et al. \cite{allen_learning_2021} developed an approach to learn Markovian abstract states by combining inverse model estimation and temporal contrastive learning. Zheng et al. \cite{zheng_intention-conditioned_2025} build a probabilistic model to predict which states an agent will visit in the future using flow matching, but it also necessitates the use of its own RL update algorithm and thus can not be used with existing prevalent RL algorithms. It is worth noting that, despite sharing a similar core motivation, these methods have not been evaluated with either/both PPO algorithm or robot locomotion learning. Adapting them to this setting would likely require substantial architectural modifications and hyperparameter tuning. Consequently we do not regard them as directly comparable baselines for our approach.

\subsection{Neural dynamics models for articulated robots locomotion} 
There has been sustained interest in modeling agent--environment dynamics and leveraging such models for control and decision-making \cite{hou2026world}. However, owing to its complexity, the literature on \textit{articulated robot locomotion} remains limited compared with the broader field of agent--environment modeling. Existing approaches can be broadly categorized into two paradigms.

\paragraph{World-model-style approaches (forward dynamics modeling)} 
World models learn to predict future states or observations conditioned on the current state and action using large-scale interaction datasets \cite{xu2025neural,li_robotic_2025}. These learned models act as neural simulators that generate autoregressive imagined rollouts that are subsequently used for policy optimization. However, the knowledge captured by the world model is transferred to the control policy only indirectly through imagined trajectories, resulting in a looser coupling between representation learning and policy learning.

\paragraph{Tracker-style approaches (inverse dynamics modeling)} 
Another paradigm formulates motion tracking as a pretraining objective, where controllers are trained to track reference trajectories in a demonstration dataset \cite{luo2024universal,luo2025sonic}. The tracking policies can then be incorporated as modules in task-specific reinforcement learning. However, rather than learning to track arbitrary feasible motions, existing methods primarily focus on human demonstration trajectories collected from carefully curated motion-capture datasets. This reliance restricts their ability to generalize to motions that are underrepresented or absent from the demonstration dataset, and limits their applicability particularly for non-humanoid robots. In contrast, our work investigates effective pretraining from unstructured interaction datasets.

\section{Preliminaries}

Robot locomotion problems are typically represented as Partially Observable Markov Decision Processes (POMDPs), where a policy $\pi: \mathcal{O} \rightarrow \mathcal{A}$ directly maps observations $\mathcal{O}$ to actions $\mathcal{A}$ and aims to maximize the cumulative reward. The reward function $R(s_t,a_t,s_{t+1})$ encodes task objectives, where $s_t,s_{t+1}\in\mathcal{S}$ are the current and next state, respectively, and $a_t\in\mathcal{A}$ is the action taken at timestep $t$. Specific to robot locomotion tasks, the observation is often the conjunction of command $\mathcal{C}$, proprioception $\mathcal{X}$, exteroception $\mathcal{X}_e$, and last action(s) $\mathcal{A}$.

In RL, a large family of actor-critic algorithms \cite{actor_critic_algo} has been widely applied in robotics, among which Proximal Policy Optimization (PPO) \cite{schulman2017proximal} is particularly prominent. These constitute an important class of RL algorithms that integrate policy optimization with value function estimation. The actor updates the policy that generates actions, while the critic estimates the value function of the current policy, thereby reducing variance and improving the stability of learning.

Existing works often parametrize both the actor and critic networks with a simple Multi-Layer Perceptron (MLP) and initialize the weights randomly \cite{he2015delving}. Due to the large variety of possible configurations and dimensions of observation and user command, pretraining a complete model for all downstream tasks is impractical without unifying the observation and command spaces for all tasks. We address this by adopting a modular architecture.

\section{Methodologies}

\subsection{Problem formulation}
\label{sec:method-problem-formulation}

For robot locomotion, we hypothesize that actor-critic algorithms converge more rapidly when the policy and value networks incorporate representations capable of solving inverse dynamics (this is supported by empirical evidence presented in Sec. \ref{sec:experiments-ablations} and \ref{sec:update_magnitude}).
We thus propose splitting the vanilla MLP structure into multiple distinct blocks (see Fig. ~\ref{fig:pidm_archi}). One of these blocks is our proposed Proprioceptive Inverse Dynamics Model (PIDM), which we define as a mapping $I(a_t \mid x_{t-K:t+1}, a_{t-K:t-1})$, where $x_t \in o_t$ denotes the proprioception at timestep $t$, $a_t$ denotes the action taken at timestep $t$, and $K$ denotes the length of the history sequence.

\subsection{Overview}
\label{sec:method-overview}

Our overall goal is to pretrain a PIDM model using supervised learning, which can later be integrated into the actor and critic networks of PPO. First, we collect proprioceptive transition data ($x_t, a_t, x_{t+1}$) in a task-agnostic manner from the RL training process of an exploration policy \cite{sekar2020planning}. It is important to note that we solely collect transitions that resemble those from the early stages of task-specific RL training, rather than task-specific expert policy rollouts. On the one hand, this design ensures that the method does not rely on prior knowledge of the downstream tasks, nor on access to a (near) expert policy. On the other hand, the state distribution of randomly initialized policies for different tasks is very similar (see Section \ref{sec:experiments-pretrain-pidm}). Therefore, the extracted knowledge should be widely utilizable. By pretraining with this data, the model encapsulates knowledge equivalent to what it would learn in the first iterations of RL (\textit{i.e.}, basic kinematics, dynamics, and stability), enabling it to specialize in learning task-specific skills faster. We integrate the core parts of our pretrained PIDM with randomly initialized MLP modules to constitute the actor and critic networks in RL (see Fig. ~\ref{fig:pidm_archi}). To account for the lack of data capturing task-specific dynamics in the pretraining dataset, we allow the PIDM module to be updated in conjunction with the added non-pretrained parts throughout the RL process.

\subsection{Exploration-based data collection}
\label{sec:method-exploration}

    \begin{figure}
        \centering
        \includegraphics[width=0.9\linewidth]{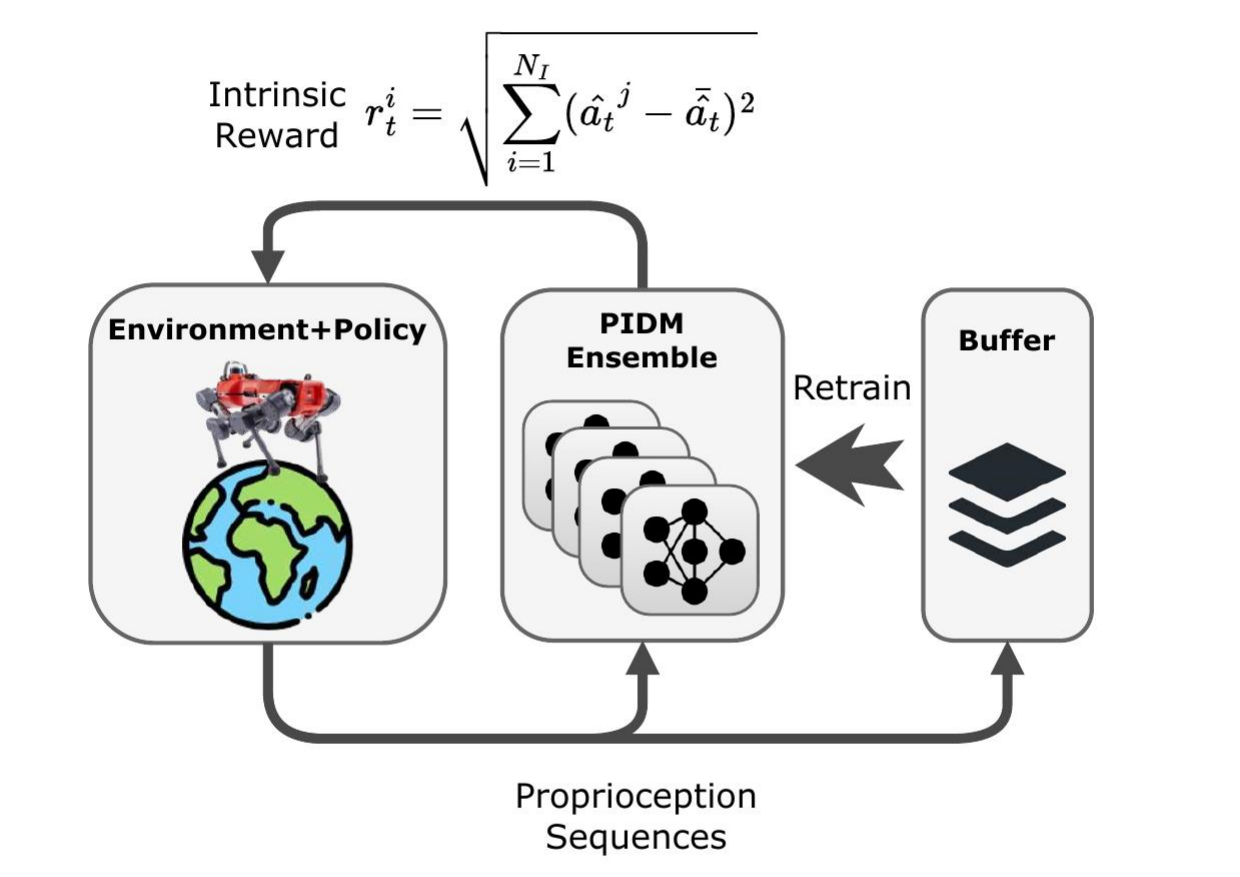}
        \caption{Diagram for exploration-based data collection pipeline, showcasing how the simulation collects data, and is guided by the ensemble of PIDM models that are periodically retrained using the buffered data.}
        \label{fig:exploration-based data collection}
    \end{figure}

We employ an exploration-based data collection strategy, inspired by \cite{pmlr-v97-pathak19a, chua_deep_2018}, outlined in Fig.  \ref{fig:exploration-based data collection}. We use it to obtain data samples that capture the jittery, exploratory behaviors commonly observed in the early stages of RL. For this purpose, we train an exploration policy with PPO, where the transitions from the on-policy rollouts are accumulated into a buffer. A probabilistic ensemble of PIDM models is frequently retrained using a bootstrap approach, where data is sampled with replacement from the buffer. The training of the exploration policy is primarily guided by the disagreement in predictions in the ensemble, as a measure of epistemic uncertainty for the PIDM inference. This incentivizes the policy to explore states where the accuracy of the PIDM can be improved with more data. 
We refrain from using the prediction error from a single PIDM model as intrinsic reward to avoid its bias toward large-magnitude actions and high-frequency jitter motions. 
Secondary rewards added include a minimal set of regularizing rewards to constrain unwanted behaviors (e.g., high action rates, torques, or joint velocities, too short foot in-the-air time) that are common to any task.
During data collection, we employ standard domain randomization techniques for RL training \cite{miki_learning_2022, lee2020learning, kaidanov2024role}, such as varying the robot link masses, the friction coefficients, and applying random pertubation forces.

\subsection{Pretraining the Proprioceptive Inverse Dynamics Model}

    \begin{figure}
        \centering
        \includegraphics[width=\linewidth]{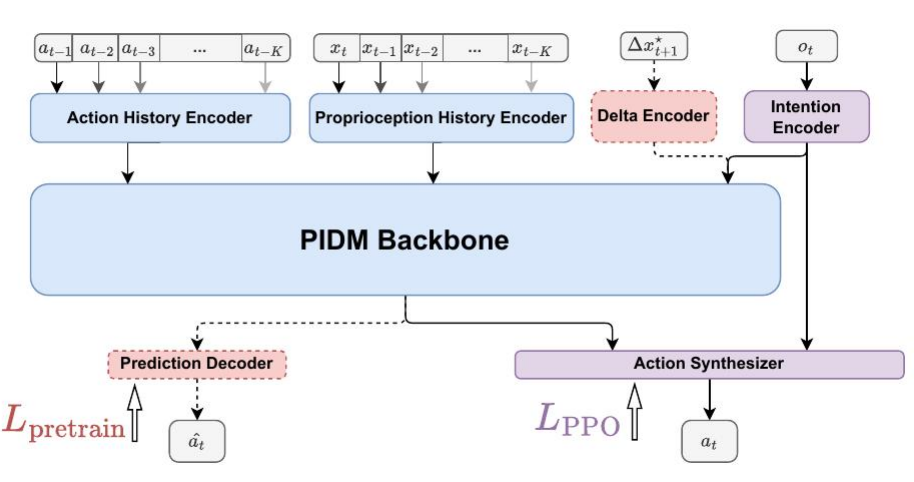}
        \caption{Proprioceptive Inverse Dynamics Model (PIDM) architecture and its integration into the actor network. During pretraining of the PIDM, the dashed \textcolor{BrickRed}{red} parts of the network are included. However, when integrating into the actor-critic structure, those are removed and replaced by the encoder and decoder in \textcolor{Thistle}{purple}.}
        \label{fig:pidm_archi}
    \end{figure}

\label{sec:method-pretrain}
We parameterize the PIDM with an MLP-based modular architecture, as shown in Fig.  \ref{fig:pidm_archi}. The model takes as input a history of actions $a_{t-K:t-1}$ and proprioceptive observations $x_{t-K:t+1}$ of length $K$. Both are passed through a dual-layer MLP encoder before being fed into the \emph{PIDM backbone}, which is a 4-layer MLP. During pretraining, we specify a desired delta-state $\Delta x_{t+1}^*$ to achieve in the next time step, and use an $L1$ loss to supervise the PIDM to output the required action $a_t$ to reach the corresponding target future state $x_{t+1}^*$. The pretraining dataset is also augmented with symmetry transformations, as defined by \cite{mittal2024symmetry} or \cite{byun2024symmetric}, and observation noise to improve robustness and increase sample diversity. 

The necessity of including a history of proprioception for PIDM is mainly due to the fact that terrain information and contact states are impossible to deduce from one single frame of proprioception, which is further worsened by the noise and domain randomization in the training process. However the action history together with proprioception history can provide indirect observability to these crucial clues about the system's dynamics.
It is also important to note that the PIDM model does not rely on any privileged information input. 

\subsection{Warm-starting Reinforcement Learning}

\subsubsection{Integrating PIDM into actor-critic networks} The pretrained PIDM is integrated into both the actor and critic networks. As shown in Fig.  \ref{fig:pidm_archi}, for the actor, we first remove the \textit{Delta Encoder} and substitute it with a randomly initialized \emph{Intention Encoder} that processes the complete task-specific observation. The \emph{Intention Encoder} now only needs to learn an embedding-based representation of the task-specific delta target state $\Delta x_{t+1}^*$, which can be preprocessed by the pretrained \emph{PIDM Backbone}. Meanwhile, the original \emph{Prediction Decoder} is removed, and the concatenated outputs of the \emph{PIDM Backbone} and \emph{Intention Encoder} are passed in to a randomly initialized \emph{Action Synthesizer} that synthesizes the final action $a_t$. PIDM is used in the critic via an almost identical architecture, with the only difference that the \textit{Action Synthesizer} in the actor is replaced with a \textit{Value Synthesizer} that outputs a scalar value estimation. All aforementioned submodules are instantialized as MLPs.

The addition of the \emph{Intention Encoder} is necessary to ensure dimension compatibility and enable the training to steer the pretrained module. The task-specific observation $o_t$ can be anything and is totally independent of our proposed approach. We also empirically discovered that the inclusion of the randomly initialized \emph{Action Synthesizer} is crucial for stabilizing the training by ensuring that the action mean at the beginning of RL is almost 0 given any input, similarly to the case with a randomly initialized vanilla MLP. 

\subsubsection{Framework compatibility} Except for the architectures of the actor and critic networks and the way the weights are initialized, our method does not require any modifications to either the POMDP (reward, curriculum design, observations, actions, and terminations) or to the PPO update rules and hyperparameters. The task-dependent \emph{Intention Encoder} and \emph{Action Synthesizer} can adapt to any configuration and dimension of the input and output. Therefore, the feasibility of handling arbitrary tasks is not limited. Every parameter in the pretrained PIDM remains trainable during the RL process.

\section{Experiments}

\begin{figure*}[thb!]
    \centering

    \begin{subfigure}[t]{0.28\textwidth}
        \centering
        \includegraphics[width=\linewidth]{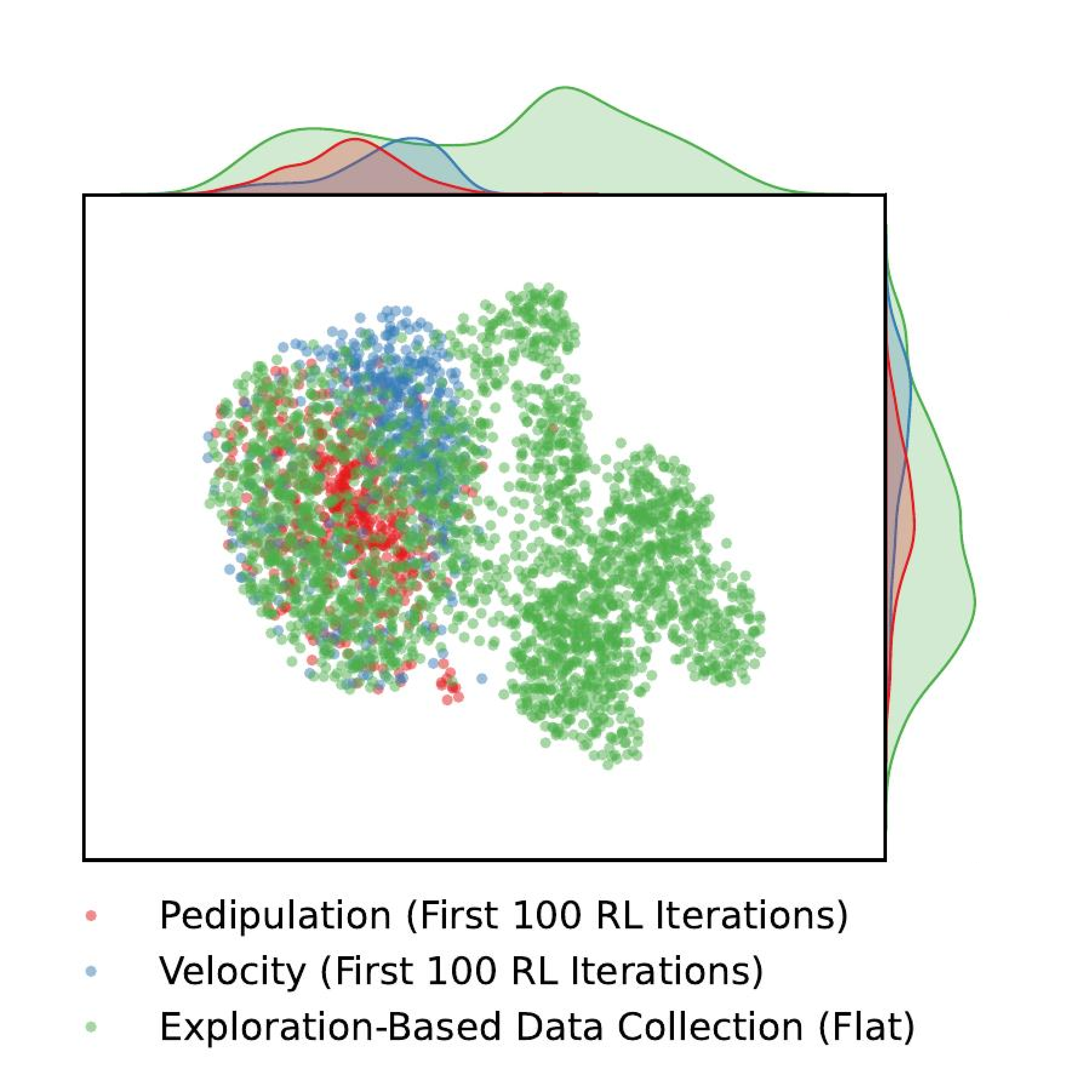}
        \caption{Training data projection}
        \label{fig:umap_data_coverage}
    \end{subfigure}
    \hfill
    \begin{subfigure}[t]{0.35\textwidth}
        \centering
        \includegraphics[width=\linewidth]{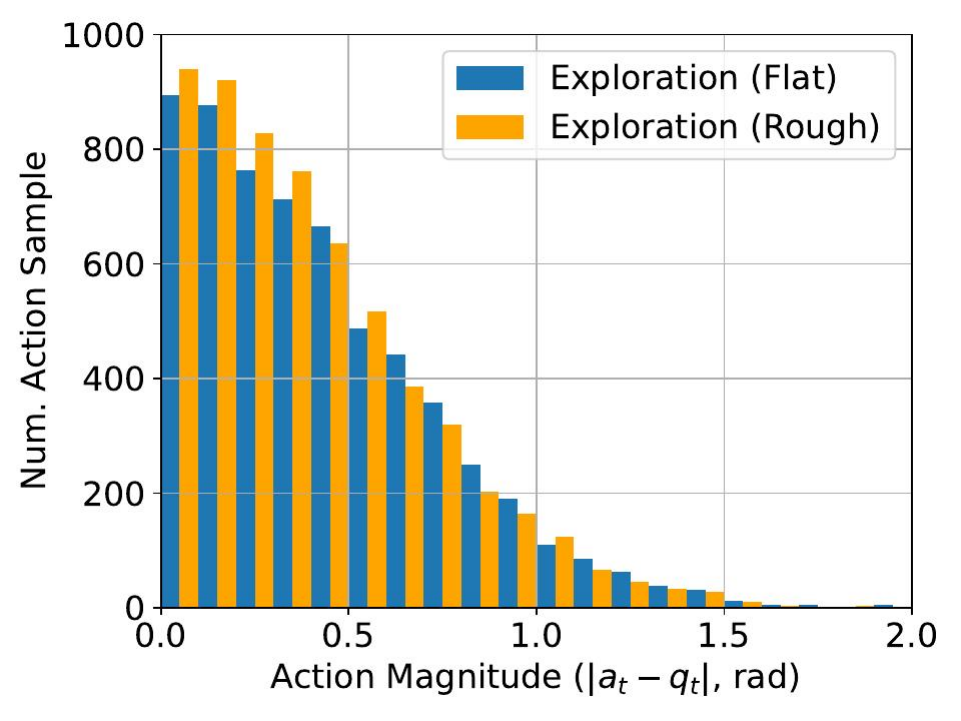}
        \caption{Sample distribution}
        \label{fig:data-distribution}
    \end{subfigure}
    \hfill
    \begin{subfigure}[t]{0.35\textwidth}
        \centering
        \includegraphics[width=\linewidth]{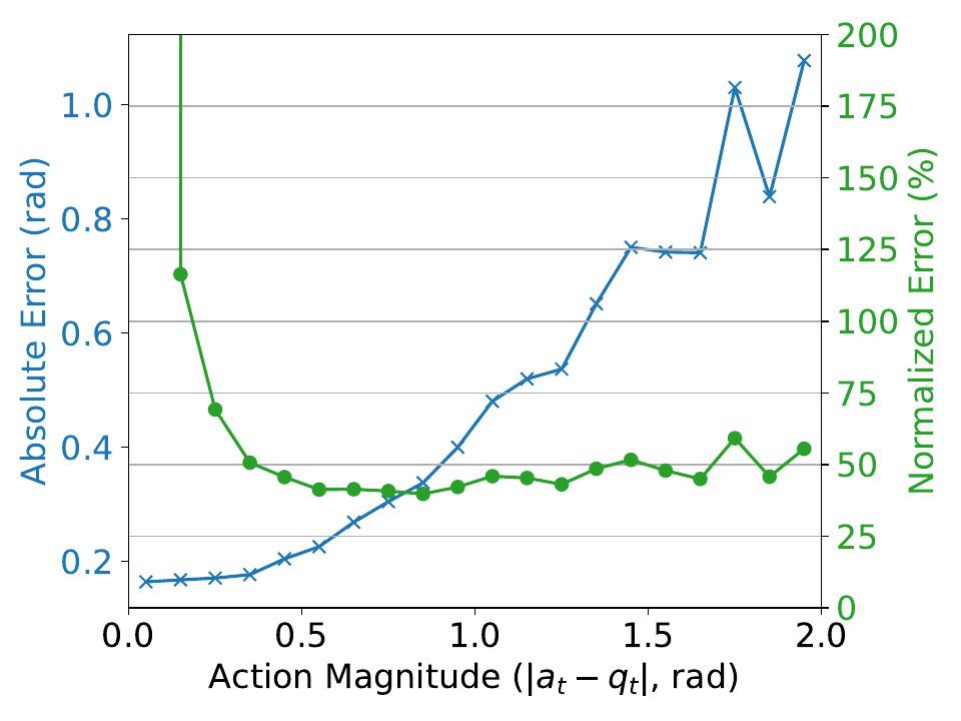}
        \caption{PIDM prediction accuracy}
        \label{fig:pidm-accuracy}
    \end{subfigure}
    
    \caption{\textbf{PIDM training and dataset analysis of ANYmal D:} For the pretraining dataset we visualize (a) its coverage (green, task-agnostic) compared to the initial exploration stages in RL (red and blue, task-specific) using an UMAP projection, and (b) its sample distribution of absolute action magnitudes $|a_t - q_t|$ over different terrains. Finally, in (c) we show the resulting PIDM accuracy across the entire action range as absolute joint errors $|\hat{q}_{t+1} - q_{t+1}|$ and  normalized by the action magnitude.}
    \label{fig:PIDM_accuracy}
\end{figure*}

\subsection{Reinforcement Learning Tasks}
\label{sec:tasks}

We test our method on 9 RL environments with 3 different embodiments: a) 2 blind tasks (velocity-tracking locomotion \cite{rudin2022learning} and pedipulation \cite{arm_pedipulate_2024}) and 5 perceptive tasks (parkour walk, climb up, climb down, crouch, and jump \cite{hoeller_anymal_2023}) with ANYmal-D, b) velocity-tracking locomotion task with Unitree Go1 quadrupedal robot (default implementation in \cite{mittal2025isaac}), and c) velocity-tracking locomotion task with Unitree G1 humanoid robot (default implementation in \cite{mittal2025isaac}). All training is performed in Isaac Lab \cite{mittal2025isaac} with a customized fork of open-source RL library RSL-RL \cite{schwarke2025rslrl}.  

Despite the diverse rewards, curricula and hyperparameters involved in aforementioned works, the network architectures used in the original implementations are very similar: the actor and critic networks are both 4-layer MLPs. 
Our proposed architecture has approximately {4$\times$} the number of parameters due to the inclusion of state history and the need to cover a larger initial state space in pretraining, whilst in comparison task-specific policies can immediately hyper-specialize.

\subsection{Pretraining the Proprioceptive Inverse Dynamics Model}
\label{sec:experiments-pretrain-pidm}

In this subsection, we describe how to obtain a pretrained PIDM model and analyze both the dataset distributions and the model’s accuracy, using ANYmal-D as an example. We first analyze the quality of the data collected using the exploration-based strategy outlined in Section \ref{sec:method-exploration}. In addition to the previously mentioned standard data augmentations (\textit{e.g.}, mass randomization, random noise, symmetry), we collect data on either or both flat and basic rough terrain generated with Perlin noise \cite{miki_learning_2022, lee2020learning}. In Fig. ~\ref{fig:umap_data_coverage} we plot samples from the flat-terrain environments along with samples from the learning process of  \textit{Pedipulation} and \textit{Locomotion} tasks, which are trained solely on flat terrain. Using UMAP \cite{mcinnes2018umap}, we project the proprioceptive observations $x$ for our collected dataset into 2D, and the observations from the first $100$ iterations of RL training for pedipulation and locomotion. We can thus validate that we obtain good coverage of, and beyond, the initial stages of the RL training process, which aligns with the goals outlined in Section~\ref{sec:method-overview}.

The PIDM is pretrained as described in Section \ref{sec:method-pretrain}. For each embodiment, we use a total of 7 million samples for training and a separate validation set. For plotting purpose only, we randomly select $1,000$ validation samples and consider only the $12$ joint angles $q$ of ANYmal-D. Fig. ~\ref{fig:data-distribution} shows the distribution of the action magnitude, \textit{i.e.} the magnitude in radians of the commanded changes in joint angles. Fig. ~\ref{fig:pidm-accuracy} shows the final prediction accuracy of the trained PIDM. We show both the absolute error and the normalized error, which is the error expressed as a fraction of the action magnitude. It achieves a normalized error of around $40\% \sim 50\%$, with a minimum error of $\sim 0.1$ radians for small actions. Fig. \ref{fig:pidm-accuracy}\ together with Fig. \ref{fig:data-distribution} indicate that most of the actions are smaller than 1.0 rad and has an absolute error smaller than 0.4 rad. 

While the model is able to generate predictions in close proximity to the ground truth, the limitations in accuracy also underscore the significant difficulty of achieving high-accuracy PIDM training, which can be attributed to the large transition space, partial observability, and the lack of inductive bias inherent to MLPs. While accurate data-driven modeling of robot locomotion may be achievable with models that are many orders of magnitude larger \cite{xu2025neural} than those used here, actor and critic networks in motion-policy learning have traditionally been extremely lightweight.
The compactness of the architectures can be attributed to the fact that the trained policy networks are expected to be deployed on real mobile hardware and be reactive at a high frequency (typically $50\sim200$ Hz). 
Moreover, large models are known to make reinforcement learning substantially more challenging \cite{ota2021training, li2023survey}. As a result, significantly increasing model capacity raises concerns about whether existing methods can still be applied without modification. For these reasons, we choose to keep the model size to millions of parameters which is much closer to that of the vanilla MLPs used in prior works. Although they may not seem too accurate for an inverse-dynamics solving, we will demonstrate in the following subsection that a pretrained module of such accuracy can already significantly enhance RL training.

\subsection{Quantitative experiments}

\begin{table*}[ht!]
    \vspace{2em}
    \centering
    \caption{Increase in performance (based on reward/curriculum progress) and sample efficiency (number of iterations required to reach $90$\% of the maximum performance). Percentage values are \textit{w.r.t.} a randomly initialized PIDM model. Values are averaged across five runs with different starting seeds. We also report the performance of the 4-layer vanilla MLP for reference.}
    \resizebox{\textwidth}{!}{%
        \begin{tabular}{cccccccccc|c|c|c}
        \toprule
        \multicolumn{2}{c}{\multirow{3}{*}{\textbf{Metric}}} & \multirow{3}{*}{\textbf{Method}} & \multicolumn{7}{c|}{\textbf{ANYmal D}} & \textbf{Go1} & \textbf{G1} & \multirow{3}{*}{\textbf{Avg.}} \\ 
        
        & & & Loco- & Pedipu- & Parkour & Climb & Climb & \multirow{2}{*}{Crouch} & \multirow{2}{*}{Jump} & Loco- & Loco- & \\ 

        & & & motion & lation & Walk & Up & Down & & &motion&motion &\\ 
        
        \toprule 
        
        Final perf. & \multirow{2}{*}{(\%, $\uparrow$)} & Vanilla MLP & \good{+0.5} & \good{+0.2} & \bad{-0.8} & \bad{-3.7} & \bad{-0.3} & \best{+1.8} & \best{+11.1} & \good{+0.6}  & \bad{-0.2} & \good{+1.0} \\
        
        increase &  & PIDM (Pretrained) & \best{+10.1} & \best{+6.3} & \best{+0.7} & \best{+0.4} & {0.0} & \best{+1.8} & \good{+5.9} & \best{+3.6} & \best{+10.0} & \best{+4.3} \\ \midrule
        
        Num. iters. & \multirow{2}{*}{(\%, $\downarrow$)} & Vanilla MLP & \good{-28.7} & \bad{+5.0} & \good{-18.7} & \bad{+48.6} & \bad{+35.3} & \good{-29.5} & \best{-53.0} & \bad{+2.2} & \best{-46.7} & \good{-9.5} \\
        
        to converge & \quad & PIDM (Pretrained) & \best{-33.1} & \best{-42.0} & \best{-35.3} & \best{-15.9} & \best{-41.8} & \best{-57.3} & \good{-49.3} & \best{-17.7} & \good{-34.0} & \best{-36.2} \\ \bottomrule
        \end{tabular}%
    }
    
    \label{tab:quant}
\end{table*}

\begin{figure*}[h]
    \centering 
    \vspace{-1em}
    \includegraphics[width=0.8\textwidth]{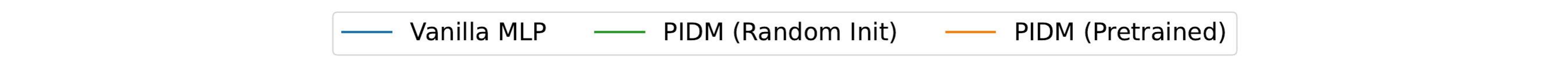}
    \vspace{0.5em}
    
    \begin{minipage}{0.24\textwidth}
        \centering
        \includegraphics[width=\linewidth]{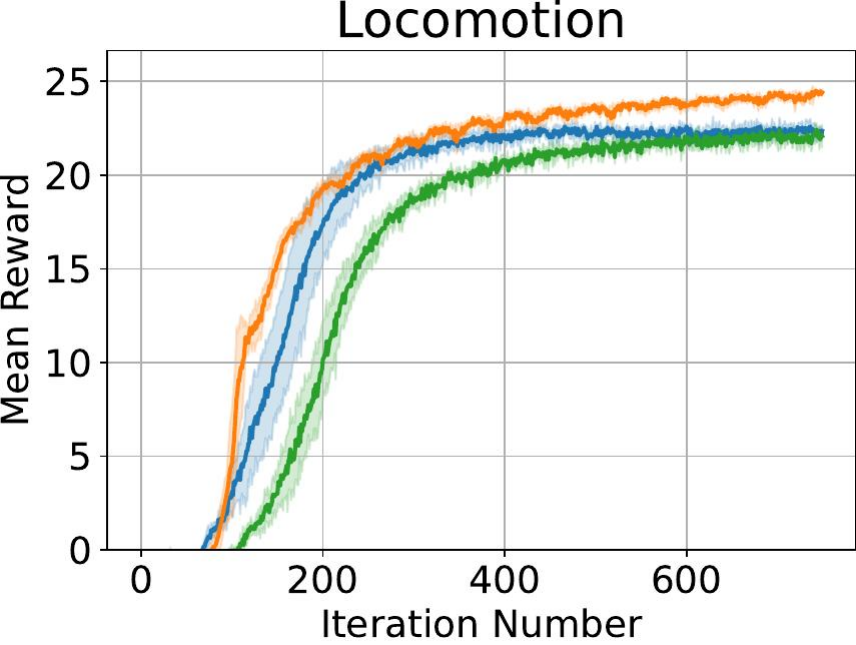}
    \end{minipage}%
    \hfill 
    \begin{minipage}{0.24\textwidth}
        \centering
        \includegraphics[width=\linewidth]{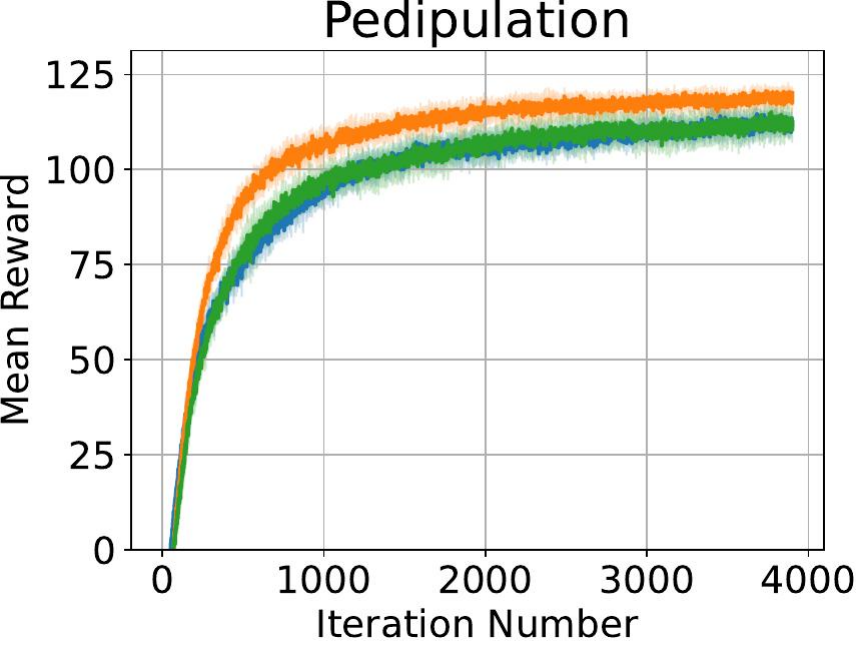}
    \end{minipage}%
    \hfill
    \begin{minipage}{0.24\textwidth}
        \centering
        \includegraphics[width=\linewidth]{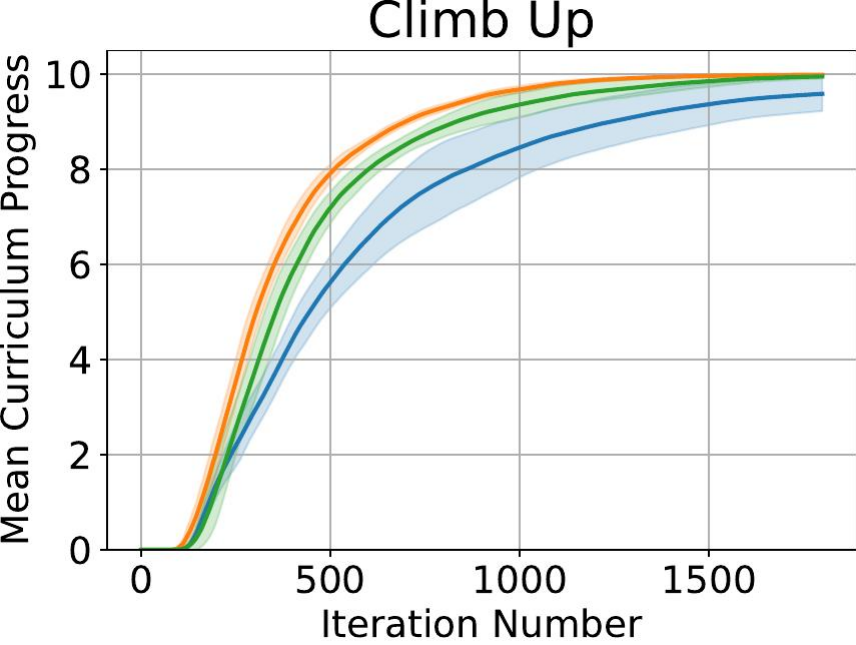}
    \end{minipage}%
    \hfill
    \begin{minipage}{0.24\textwidth}
        \centering
        \includegraphics[width=\linewidth]{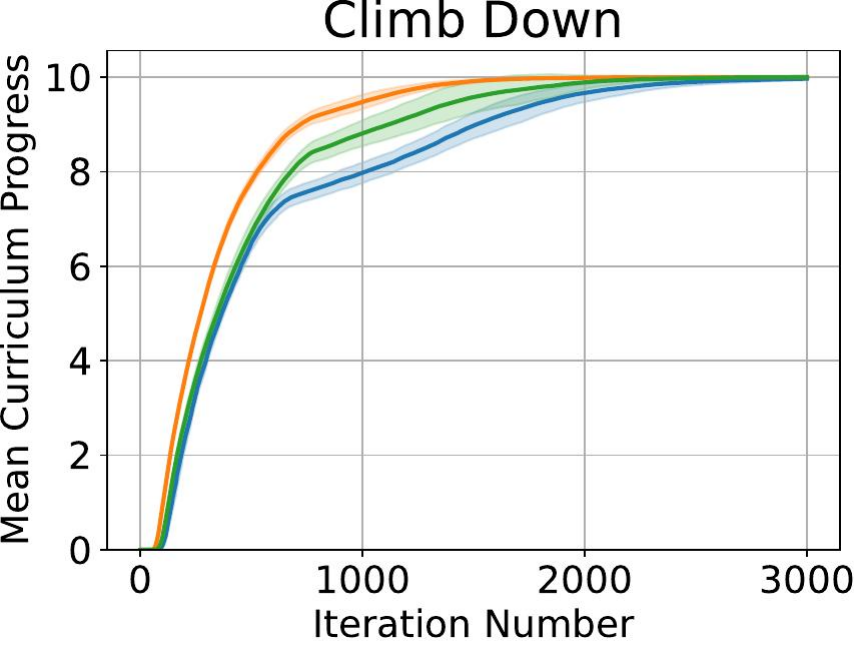}
    \end{minipage}

    \caption{Evolution of the main performance metric during training for \textit{Locomotion}, \textit{Pedipulation}, \textit{Climb Up} and \textit{Climb Down} tasks with ANYmal-D. The shaded areas denote standard deviations across five seeds.}
    \label{fig:quant_res}

\end{figure*}

In each experiment, we compare three methods: (i) the vanilla 4-layer MLP, (ii) our PIDM architecture with randomly initialized weights, and (iii) our PIDM architecture with pretrained weights. The utility of our method (\textit{i.e.} using pretrained weights) is better indicated by the comparison between (ii) and (iii), and the performance of the vanilla 4-layer MLP is included only as a reference. Results are averaged over five runs with different random seeds.
To note is that we did \emph{not} tune the learning parameters (learning rate, entropy coefficient, etc.) of the tasks, which were chosen for optimal performance of the original vanilla MLP. We merely used our architecture as a drop-in replacement. We find that incorporating a moderate degree of action symmetry loss improves training stability considerably.

For the curriculum-free tasks (pedipulation, locomotion, we use the \textit{Mean Reward} as the performance indicator. While since the adaptive (progress-based) terrain difficulty curriculum exist in all five parkour tasks, the mean reward curves can not be directly taken as performance metric of policies because of the different terrain conditions. Therefore, we use the \emph{Curriculum Progress}, indicated by the average of maximum terrain difficulty reached over all sub-environments as the performance indicator. 
We introduce two core metrics based on the performance indicator: 
\begin{itemize}
    \item \textbf{Final performance increase} expresses the percentage of change in the main performance indicator at the end of RL training compared to that of the \textit{PIDM (Random Init)} baseline.
    \item \textbf{Number of iterations to converge} is a measure of sample efficiency. This term represents the percentage of change in the number of iterations required to reach 90\% of the final performance of the \textit{PIDM (Random Init)} baseline in the main performance indicator.
\end{itemize}

The results across all nine tasks are presented in Table~\ref{tab:quant}. We also plot the evolution of the main performance indicator during training of a subset of tasks in Fig.~\ref{fig:quant_res}. The PIDM architecture with random weight initialization, \textit{PIDM (Random Init)}, generally lags behind the vanilla MLPs due to a larger model size and input dimension. However, with the proposed pretraining strategy, \textit{PIDM (Pretrained)} not only consistently outperforms \textit{PIDM (Random Init)} in all metrics, but also significantly surpasses the performance of the vanilla MLP in 7 out of 9 tasks. We speculate that matching the performance of the MLP with our PIDM architecture for the other 2 tasks can be potentially achieved with hyperparameters tuning. 

The main finding is obtained from the comparison between the randomly initialized and pretrained architectures. 
When compared with \textit{PIDM (Random Init)}, the proposed \textit{PIDM (Pretrained)} showcases an improvement of \overallperfimpro \space on final performance, and enhances sample efficiency by a margin of \overallseimpro. We also note that despite the PIDM never having experienced the complex terrains used in the parkour task, it quickly adapts to the new task-specific dynamics during RL training.

\subsection{Ablations}
\label{sec:experiments-ablations}

\begin{table*}[ht!]
    \centering

    \begin{minipage}[c]{0.58\textwidth}
        \centering
        \captionof{table}{Ablation on using pretrained weights to initialize either the actor, critic, or both. Results are in comparison to a fully randomly initialized PIDM architecture.}
        \resizebox{\linewidth}{!}{
        \begin{tabular}{c|c|cc}
            \toprule
            \multirow{3}{*}{\textbf{Metric}} & \multirow{3}{*}{\textbf{Method}} & \multicolumn{2}{c}{\textbf{Anymal D}} \\
            
             &  & Climb & Climb \\ 
             & & Up & Down \\ \toprule 
            \multirow{3}{*}{\makecell{Final perf. \\ increase (\%, $\uparrow$)}}
            & PIDM (Pretrained Actor Only) & 0.0 & 0.0\\ 
            & PIDM (Pretrained Critic Only) & \good{+0.2} & 0.0\\ 
            & PIDM (Pretrained Both) & \best{+0.4} & 0.0\\ \midrule
            
            \multirow{3}{*}{\makecell{Num. iters. to \\ converge (\%, $\downarrow$)}}
            & PIDM (Pretrained Actor Only) & \good{-10.9} & \good{-34.7} \\ 
            & PIDM (Pretrained Critic Only) & \good{-1.1} & \good{-21.5} \\ 
            & PIDM (Pretrained Both) & \best{-15.9} & \best{-41.8}\\  
            \bottomrule
        \end{tabular}
        }
        
        \label{tab:actor_critic_ablation}
    \end{minipage}
    \hfill
    \begin{minipage}[c]{0.4\textwidth}
        \centering
        \includegraphics[width=\linewidth]{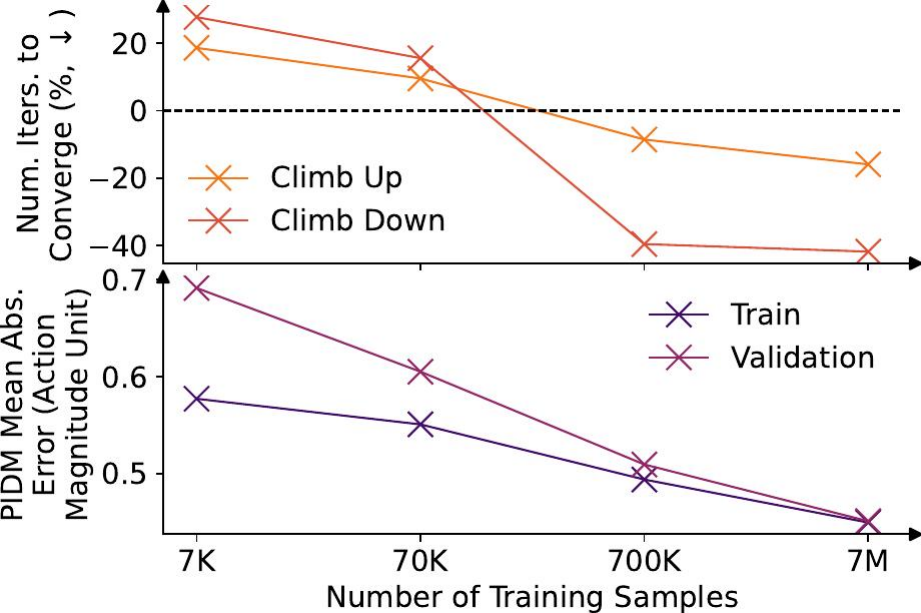}
        \captionof{figure}{PIDM mean absolute errors and relative convergence speed on downstream RL tasks across different pretraining dataset sizes.}
        \label{fig:num_train_samples_ablation}
    \end{minipage}
\end{table*}

\begin{figure*}[ht!]
    \centering
    \includegraphics[width=1.0\linewidth]{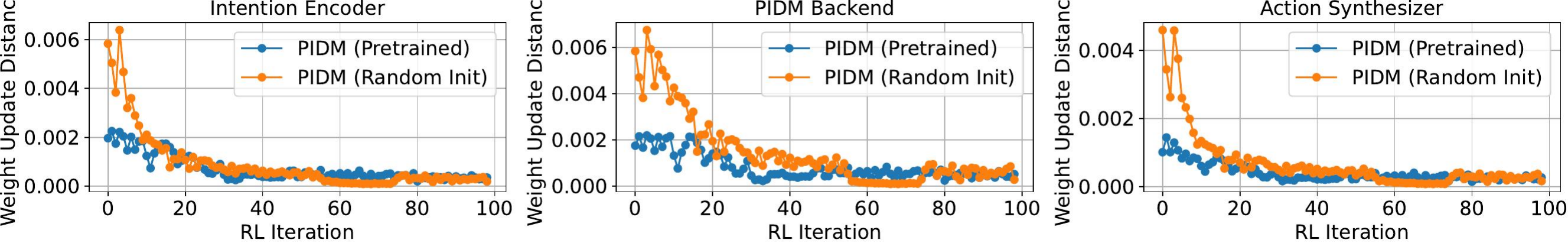}
    \caption{Network weight update magnitude comparison in the PIDM structured actor network during ANYmal-D pedipulation RL training. In each submodule, the update of each linear layer weight is indicated by the mean absolute change per parameter, which is then averaged over all layers.}
    \label{fig:weight_update_magnitude}
\end{figure*}

We also perform 2 ablations with \textit{Climb Up} and \textit{Climb Down} tasks of ANYmal-D, to motivate some of our design choices. 

First, in Table~\ref{tab:actor_critic_ablation} we analyze the initialization strategy of choosing to pretrain either or both of the actor and critic networks. We find that pretraining both is the best approach as expected, and that only pretraining either the actor or critic generally still improves mean performance.

Secondly, we ablate the size of training set used in pretraining. We respectively pretrained the PIDM model with 7k, 70k, 700k and 7M (main proposed size) samples drawn from the exploration dataset for 100 epochs, and then validated them on the same separate validation set containing 600k samples. The pretrained weights were subsequently loaded into RL networks of the 2 tasks to compare their learning outcome. Their training losses and validation losses are plotted in Fig. \ref{fig:num_train_samples_ablation}. The results show a monotonic increase in learning performance as the training set grows. Notably, pretraining with too few samples (7k or 70k) underperforms relative to random initialization, likely because overfitting produces non-generalizable representations. Given this trend, we suppose more performance gain is possible with advanced model architecture that is able to model complex distributions and achieve higher accuracy.

\subsection{Weight Update Magnitude} 
\label{sec:update_magnitude}

We compare the network weight update magnitudes between \textit{PIDM (Pretrained)} and \textit{PIDM (Random Init)} during the first 100 iterations of RL in Fig.  \ref{fig:weight_update_magnitude}. We find that not only does the model exhibit smaller updates per iteration in the pretrained \emph{PIDM backbone}, but this also results in smaller updates in the randomly initialized upstream \emph{Intention Encoder} and downstream \emph{Action Synthesizer}. This finding suggests that our pretrained weights lie closer to the desired local minimum and is an indicator that the optimization process can properly leverage this fact. 


\section{Conclusion}
To summarize, we have presented a method for warm-starting the RL training process in actor-critic algorithms, targeted for robotic motion control. Our proposed approach leverages a network architecture based on a Proprioceptive Inverse Dynamics Model (PIDM) that is pretrained using exploration-based data from a specific robot embodiment. Our modular architecture functions as a drop-in replacement without hyperparameter tuning, facilitating learning diverse tasks with the robot embodiment of pretraining. We demonstrate on 9 diverse RL environments with 2 quadrupedal robots and 1 humanoid robot that we can improve the final performance by \overallperfimpro, and enhance sample efficiency by \overallseimpro. Our approach is generic and consistently outperforms random initialization when trained on sufficiently large pretraining datasets, making it a versatile and broadly applicable technique. We also provide ablation studies to motivate our design choices and extensive empirical insights into the inner workings of our method. Future work will focus on incorporating articulation-aware inductive biases into architectural design, as well as progressively leveraging new data to finetune pretrained PIDM, thereby better supporting continuous learning.





\bibliographystyle{IEEEtran}
\bibliography{IEEEabrv, main}


\end{document}